\documentclass[10pt,twocolumn,letterpaper]{article}
\usepackage{cvpr}
\usepackage{times}
\usepackage{epsfig}
\usepackage{graphicx}
\usepackage{amsmath}
\usepackage{amssymb}
\usepackage{enumitem}
\usepackage{color}
\usepackage{here}
\usepackage{amsmath}
\DeclareMathOperator*{\argmax}{arg\,max}

\usepackage[normalem]{ulem}

\usepackage{caption}
\usepackage{subcaption}
\usepackage{authblk}
\usepackage{lipsum}
\usepackage{algorithm}
\usepackage{algpseudocode}
\usepackage{url}

\newcommand*{\affaddr}[1]{#1} 
\newcommand*{\affmark}[1][*]{\textsuperscript{#1}}

\newcommand\blfootnote[1]{%
  \begingroup
  \renewcommand\thefootnote{}\footnote{#1}%
  \addtocounter{footnote}{-1}%
  \endgroup
}

\usepackage[pagebackref=true,breaklinks=true,letterpaper=true,colorlinks,bookmarks=false]{hyperref}

\cvprfinalcopy 

\def\cvprPaperID{4606} 

\ifcvprfinal\fi
\begin{document}

\title{Learning to Explain with Complemental Examples}

\author{%
 Atsushi Kanehira\affmark[1] and Tatsuya Harada\affmark[2,3]\\
\affaddr{\affmark[1]Preferred Networks},
\affaddr{\affmark[2]The University of Tokyo}, \affaddr{\affmark[3]RIKEN}}


\maketitle
\thispagestyle{empty}

\begin{abstract}
This paper addresses the generation of explanations with visual examples. Given an input sample, we build a system that not only classifies it to a specific category, but also outputs linguistic explanations and a set of visual examples that render the decision interpretable. Focusing especially on the complementarity of the multimodal information, i.e., linguistic and visual examples, we attempt to achieve it by maximizing the interaction information, which provides a natural definition of complementarity from an information theoretical viewpoint. We propose a novel framework to generate complemental explanations, on which 
the joint distribution of the variables to explain, and those to be explained is parameterized by three different neural networks: predictor, linguistic explainer, and example selector. Explanation models are trained collaboratively to maximize the interaction information to ensure the generated explanation are complemental to each other for the target. The results of experiments conducted on several datasets demonstrate the effectiveness of the proposed method.
\end{abstract} 

\section{Introduction}
\blfootnote{This work is done at the University of Tokyo.}
When we explain something to others, we often provide supporting examples. This is primarily because examples enable a concrete understanding of abstract explanations. With regard to machines, which are often required to justify their decision, do examples also help explanations?

\begin{figure}[t!]
\vspace{-0.4cm}
  \includegraphics[width=\linewidth, height=6cm]{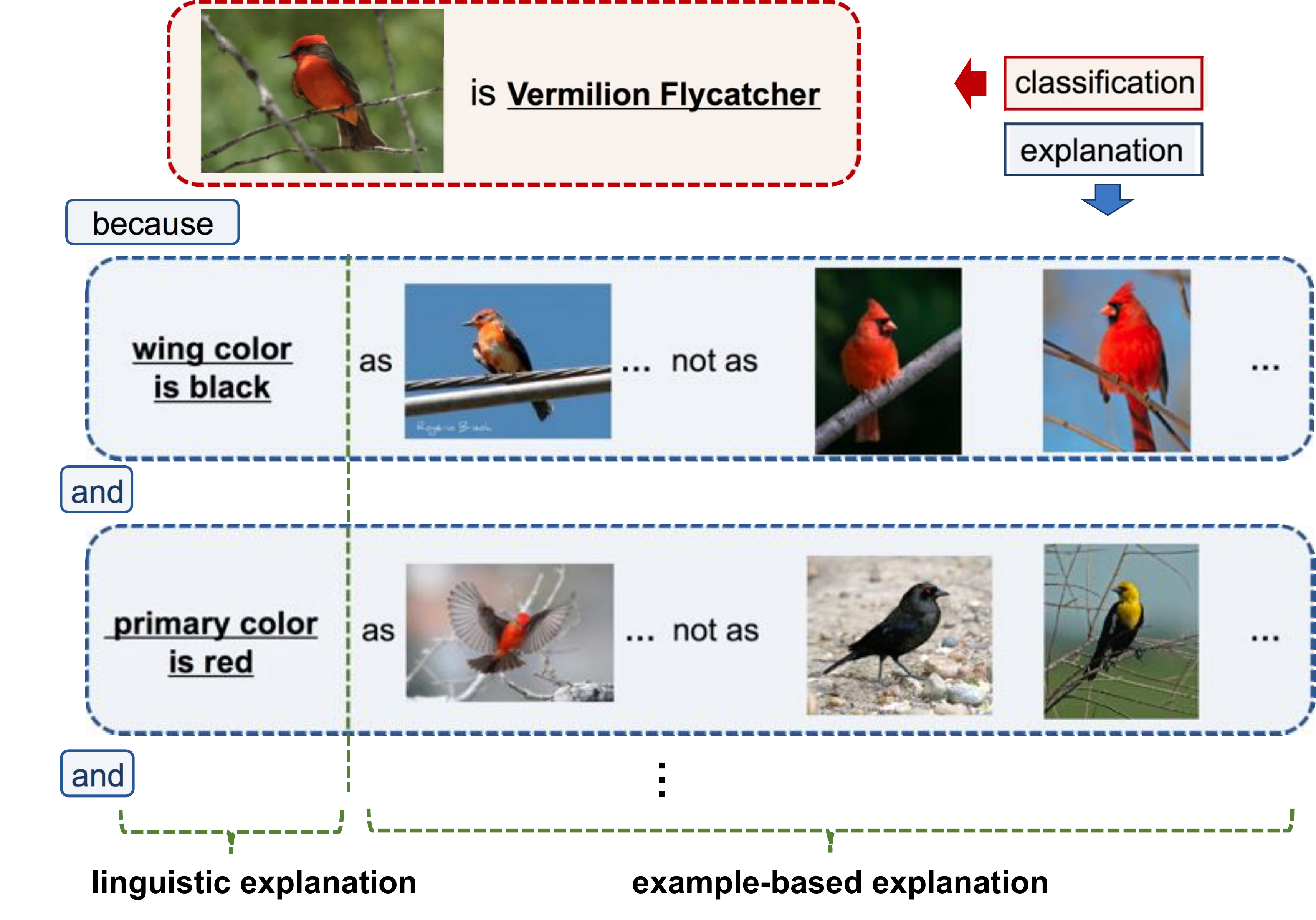}
  \caption{Our system not only classifies a given sample to a specific category (in the red dotted box), but also outputs linguistic explanations and a set of examples (in the blue dotted box).}
  \label{fig:withexample}
  \vspace{-0.6cm}
\end{figure}

This paper addresses the generation of visual explanations with visual examples. More specifically, given an input sample, we build a system that not only classifies it to a specific category but also outputs linguistic explanations and a set of examples that render the decision interpretable. An example output is shown in Fig.~\ref{fig:withexample}.

The first question to be raised toward this problem would be
``\textbf{How do examples help explanations?}'', or equivalently, ``\textbf{Why are examples required for explanations?}''\\
To answer these questions, we consider the characteristics of two types of explanations pertaining to this work: linguistic explanation, and example-based explanation. 

\begin{itemize}[noitemsep,topsep=0pt, leftmargin=5.5mm]
\setlength{\itemsep}{0pt}
  \setlength{\parskip}{0pt}
\item Using language, one can transmit information efficiently by converting an event to a shared concept between humans. Inherently, 
the conversion process is invertible; thus, the whole event can not necessarily be represented by language alone.

\item Using examples, one can transmit information more concretely than language can, as the saying, ``a picture is worth a thousand words.'' However, the way of the interpretation for the given examples is not determined uniquely. Thus, using examples alone is inappropriate for the explanation.
\end{itemize}
These explanations with different characteristics can be expected to complement each other, that is, from a lexicon, {\it a thing that contributes extra features to something else in such a way as to improve or emphasize its quality}~\cite{complement}.

The next important questions here are as follows:
\textbf{``How can the complementarity be achieved?''} and \textbf{``Which explanation is complemental, and which is not?''}.

We answer the former question from the information theoretical viewpoint, that is, interaction-information~\cite{mcgill1954multivariate} maximization. Interaction-information is one of the generalizations of mutual information defined on more than three random variables, and provides the natural definition of the complementarity: 
The increase of dependency of two variables when the third variable is conditioned. 

We propose a novel framework in this work to build a system that generates complemental explanations.
First, we introduce a linguistic explainer and an example selector parameterized by different neural networks, in addition to the predictor that is the target of the explanation. These two auxiliary models are responsible for generating explanations with linguistic and examples, respectively, and they are simultaneously trained to maximize the interaction information between variables of explanations and the output of the predictor in a post-hoc manner. Because the direct optimization of interaction-information with regard to the selector is intractable owing to the number of combination of examples, we maximize the variational lower bound instead. One more additional classifier, referred to as reasoner, appears in the computation of the lower bound. Taking linguistics and example-based explanations as inputs, the reasoner attempts to predict the output of the predictor. To enable the optimization of the selector with back-propagation, we utilized a reparameterization trick that replaces the sampling process of the examples with a differentiable function. 

Under our framework, where complementarity is defined by information theory, we can understand better the complemental explanation related to the latter question. It can be mentioned that complemental examples for a linguistic explanation are a {\it discriminative set of examples}, by which one can reason to the correct conclusion with the given linguistic explanations, but cannot be achieved with different possible explanations. Complemental linguistic explanations to examples are also considered to be explanations that can construct such a set of examples.
More details will be discussed in the subsequent section. 

We conducted experiments on several datasets and demonstrated the effectiveness of the proposed method.

The contributions of this work are as follows:
\begin{itemize}[noitemsep,topsep=0pt, leftmargin=5.5mm]
\setlength{\parskip}{0.0cm}
\setlength{\itemsep}{0.0cm}
\item Propose a novel visual explanation task using linguistic and set of examples,
\item Propose a novel framework for achieving complementarity on multimodal explanations.
\item Demonstrate the effectiveness of the proposed method by quantitative and qualitative experiments.
\end{itemize}

The remainder of this paper is organized as follows. In Section~\ref{sec:relatedwork}, we discuss the related work of the visual explanation task. Further, we explain the proposed framework to achieve complemental explanations in Section~\ref{sec:method} and describe and discuss the experiments that we performed on it in Section~\ref{sec:experiment}. 
Finally, we conclude our paper in Section~\ref{sec:conclusion}.

\section{Related Work}\label{sec:relatedwork}
The visual cognitive ability of a machine has improved significantly primarily because of the recent development in deep-learning techniques. Owing to its high complexity, the decision process is inherently a black-box; therefore, many researchers have attempted to make a machine explain the reason for the decision to verify its trustability. 

The primary stream is visualizing where the classifier weighs for its prediction by assigning an importance to each element in the input space, by propagating the prediction to the input space~\cite{simonyan2013deep, bach2015pixel, zhang2016top, selvaraju2016grad, zhou2016learning, fong2017interpretable, zhou2018interpretable}, or by learning the instance-wise importance of elements~\cite{chen2018learning,dabkowski2017real,Kanehira_2019_CVPR_Multimodal} with an auxiliary model. 
As a different stream, some works trained the generative model that outputs explanations with natural language~\cite{hendricks2016generating, park2018multimodal} in a post-hoc manner.
 Although most studies are focused on single modality, our work exploits multimodal information for explanations.

Prototype selection~\cite{singh2012unsupervised, doersch2012makes, jain2013representing, kanehira2018aware, singh2012unsupervised, goyal2017making} or machine teaching~\cite{Aodha_2018_CVPR} can be considered as example-based explanations. The essential idea of these methods is to extract representative and discriminative (parts of) examples. In other words, they attempt to obtain examples that represent $p({\mathbf x} | c)$, which is the distribution of sample ${\mathbf x}$ conditioned on the category $c$. Our work is different in that we attempt to explain the black-box posterior distribution $p(c | {\mathbf x})$ such as that represented by deep CNN. Moreover, we utilized the linguistic information as well because the interpretation toward example-based explanation is not determined uniquely.

Few works have treated multimodality for explanation
\cite{park2018multimodal, Hendricks_2018_ECCV}, which is visual and linguistic. Although they provided visual information by referring to a part of the target samples, we explore the method to utilize other examples for explanation.

\section{Method}\label{sec:method}
The goal of this study is to build a model that generates linguistic and example-based explanations, which are complemental to each other. 
In this section, we describe the proposed framework. First, in subsection~\ref{subsec:form}, we formulate our novel task with the notation used throughout this paper. Subsequently, the objective function to be optimized is illustrated in subsection~\ref{subsec:objective}. 
From subsection~\ref{subsec:vl} to \ref{subsec:optimize}, we explain the details of the actual optimization process. The proposed method is discussed qualitatively in subsection~\ref{subsec:discuss}. Finally, its relation with other explanation methods is mentioned in subsection~\ref{subsec:relation}.

\begin{figure}[t]
  \includegraphics[width=\linewidth, height=4cm]{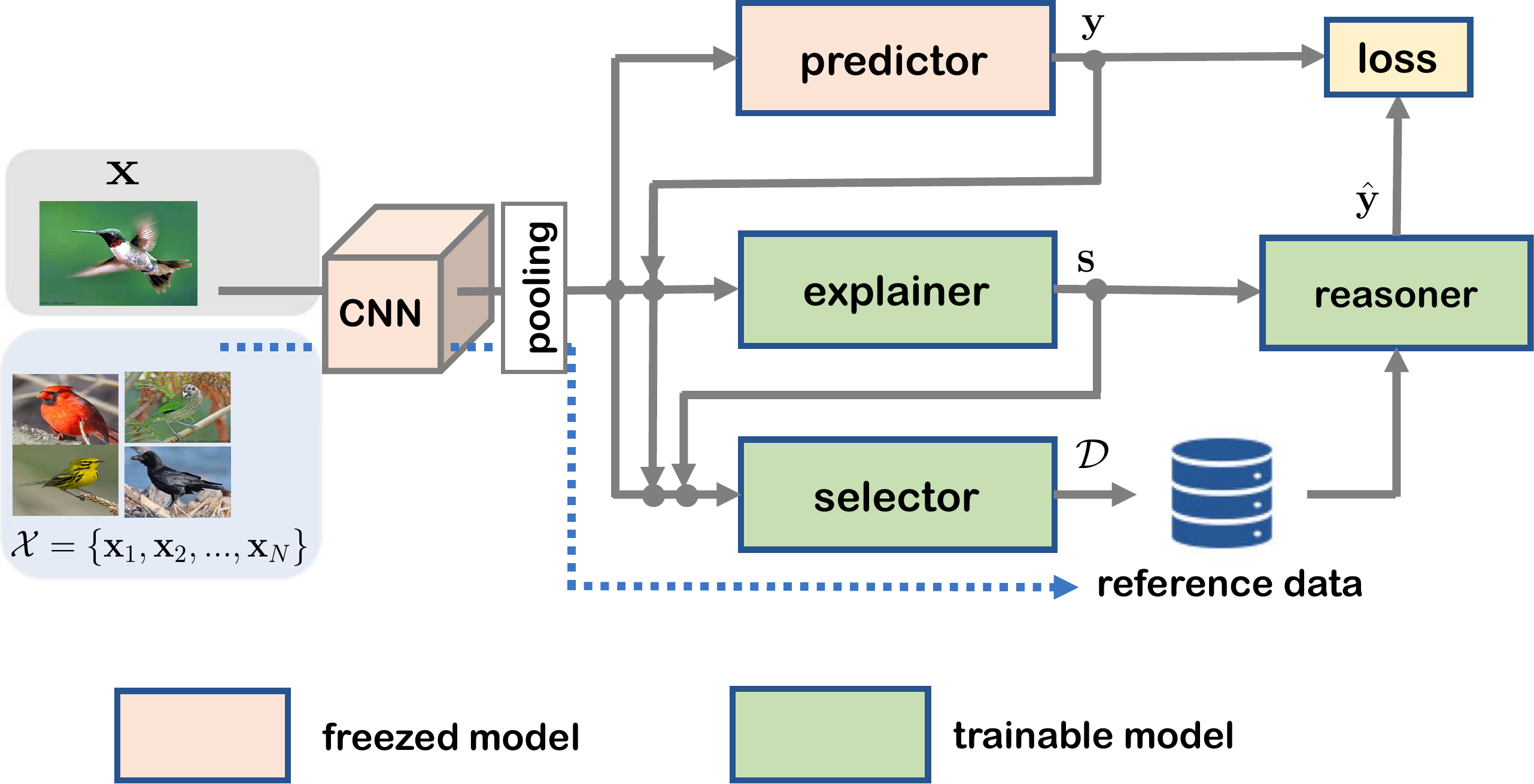}
  \caption{The pipeline of our explanation system. It holds two auxiliary models, which are responsible for generating explanations with linguistics and examples, respectively. In addition, it contains a reasoner that predicts the output of the predictor from the given explanations as described in subsection~\ref{subsec:vl}.}
  \label{fig:exp}
  \vspace{-0.5cm}
\end{figure}

\subsection{Problem formulation}\label{subsec:form}
We denote by ${\mathbf x}$ and ${\mathbf y}$ the sample and the category that are the target for explanations where ${\mathbf y}$ is a one-hot vector. ${\mathbf s}$ is a vector representing a discrete attribute, whose every index corresponds to the type of attribute~(e.g., dim1 $\rightarrow$ color, dim2 $\rightarrow$ shape..), and the value of the vector corresponds to the value of attributes~(e.g., 1 $\rightarrow$ red, 2 $\rightarrow$ blue ..). Attribute values are also treated as one-hot vector on implementation.
We assume that the attributes are assigned to all the samples used for training the explanation model. In this study, we utilize one attribute as an element of linguistic explanation. More specifically, linguistic explanation ${\mathbf s}$ contains only one non-zero value (i.e., ${||\mathbf s}||_{0}=1$), and the corresponding type-value is outputted (e.g., ``because color is red.''). To explicitly distinguish the variable representing linguistic explanation from one representing attributes assigned to the samples, we denote the former by ${\mathbf s}$ and the latter by ${\hat {\mathbf s}}$. 
The set of candidate examples used for explanation is represented by 
${\mathcal X} = \{({\mathbf x}_i, {\hat {\mathbf s}_i}, {\mathbf y}_i)\}_{i=1}^{N}$, and its subset ${\mathcal D} \subset {\mathcal X},\ |\mathcal D|=k$
is used as an element of the example-based explanation. We assume ${N}\choose{k}$, and that the number of combinations ${\mathcal D}$, is sufficiently large.
Our system generates multiple elements $({\mathbf s}_1, {\mathcal D}_1)$, $({\mathbf s}_2, {\mathcal D}_2), \ldots,({\mathbf s}_M, {\mathcal D}_M)$, and construct a explanation by simply applying them to the template as in Fig.~\ref{fig:withexample}.

We built a model not only categorizing the input ${\mathbf x}$ to a specific class ${\mathbf y}$, but also providing an explanation with linguistics and example-based explanations ${\mathbf s}$ and ${\mathcal D}$. We decomposed a joint distribution $p({\mathbf y}, {\mathbf s}, {\mathcal D}|{\mathbf x})$ to three probabilistic models: predictor, explainer, selector, all of which were parameterized by different neural networks:
\begin{align}\label{eq:decomp}
p({\mathbf y}, {\mathbf s}, {\mathcal D}| {\mathbf x}) = 
\underset{{\rm predictor}}{\underline{p({\mathbf y}| {\mathbf x})}}\ 
\underset{{\rm explainer}}{\underline{p({\mathbf s}| {\mathbf x}, {\mathbf y})}}\ 
\underset{{\rm selector}}{\underline{p({\mathcal D}| {\mathbf x},  {\mathbf y}, {\mathbf s})}}
\end{align}

\textbf{predictor} $p({\mathbf y}| {\mathbf x})$ is the target model of the explanation, which categorizes sample ${\mathbf x}$ to ${\mathbf y}$. Particularly, we study the model pretrained for the classification task.
Throughout this paper, the weight of the predictor is frozen, and the remaining two auxiliary models, namely, explainer and selector, are trained to explain the output of the predictor.
\\

\textbf{explainer} $p({\mathbf s}| {\mathbf x}, {\mathbf y})$ is the probability of linguistic explanation ${\mathbf s}$ being selected given target sample ${\mathbf x}$ and class $\mathbf{y}$. We limit ${||\mathbf s}||_{0}=1$, and the dimension and the value corresponding to the non-zero element is used as an explanation.
\\

\textbf{selector} $p({\mathcal D} | {\mathbf x}, {\mathbf y}, {\mathbf s})$ is the probability of example-based explanation ${\mathcal D}$ being selected out of all candidate examples given $\mathbf{x}$, $\mathbf{y}$, and $\mathbf{s}$ as inputs.

\subsection{Objective function}\label{subsec:objective}
We illustrate the objective function optimized for training the explanation models in this subsection. As stated earlier, linguistic explanation ${\mathbf s}$ and example-based explanation ${\mathcal D}$ are expected to be complemental to each other. Intuitively, one type of explanation should contain the information for the target ${\mathbf y}$, that is different from what the other explanation contains. 

Hence, we leverage the interaction information~\cite{mcgill1954multivariate} as an objective function. Interaction-information is a generalization of the mutual information defined on more than three random variables, and it measures how the dependency of two variable increases when the third variable is conditioned, which provides a natural definition toward complementarity.

\begin{figure*}[t]
  \includegraphics[width=\linewidth]{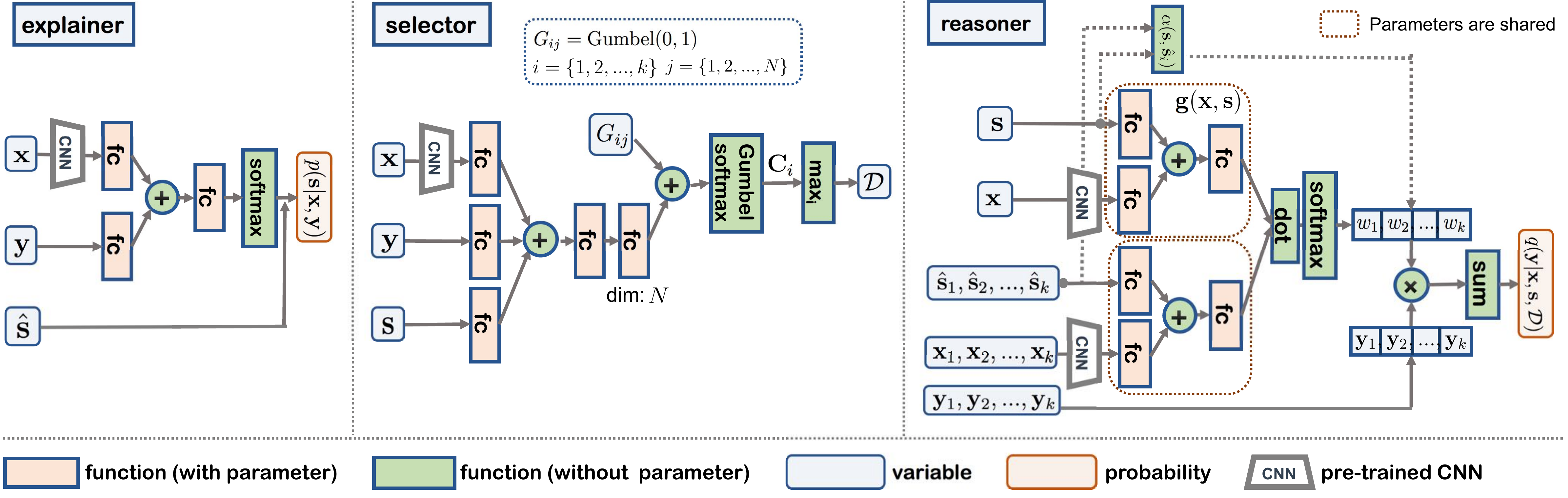}
    \vspace{-0.7cm}
  \caption{Structures of three neural networks representing three probabilistic models. As described in subsection~\ref{subsec:sampling}, the network of the selector predict the parameter of categorical distribution unlike the other two models for the ease of optimization.}
  \label{fig:structure}
  \vspace{-0.4cm}
\end{figure*}

From the definition, the interaction information of ${\mathbf y}, {\mathbf s}, {\mathcal D}$ conditioned on the input ${\mathbf x}$ is written as the difference of two mutual information:
\begin{eqnarray}\label{eq:interaction}
I({\mathbf y}, {\mathbf s}, {\mathcal D}| {\mathbf x}) = 
{I({\mathbf y}, {\mathbf s} | {\mathbf x}, {\mathcal D})} -  {I({\mathbf y}, {\mathbf s}| {\mathbf x})}  
\end{eqnarray}
where
\begin{eqnarray}\label{eq:mi1}
&&I({\mathbf y}, {\mathbf s} | {\mathbf x}, {\mathcal D})\nonumber  \\
&=&\!\!\!\!\! \int_{\mathbf x}\sum_{{\mathbf y}, {\mathbf s}, {\mathcal D}}p({\mathbf y}, {\mathbf s}, {\mathcal D} , {\mathbf x}){\rm log}\frac{p({\mathbf y}, {\mathbf s} | {\mathbf x}, {\mathcal D})}
{p({\mathbf s} | {\mathbf x}, {\mathcal D})p({\mathbf y} | {\mathbf x}, {\mathcal D})} d{\mathbf x}, \nonumber \\
&=&\!\!\!\!\!  {\mathbb E}_{p({\mathbf x})}
\underset{{\rm (A)}}
{\dashuline{\left[\sum_{{\mathbf y}, {\mathbf s}, {\mathcal D}} p({\mathbf s}, {\mathcal D} | {\mathbf x}, {\mathbf y})p({\mathbf y} |  {\mathbf x}) {\rm log} \frac{p({\mathbf s} | {\mathbf x}, {\mathbf y}, {\mathcal D})}{p({\mathbf s} |{\mathbf x}, {\mathcal D})}\right]
}}
\end{eqnarray}
and similarly,
\begin{eqnarray}\label{eq:mi2}
&&I({\mathbf y}, {\mathbf s}| {\mathbf x})\nonumber\\
&=& {\mathbb E}_{p({\mathbf x})}
\underset{{\rm (B)}}{\dashuline{
\left[ 
\sum_{{\mathbf y}, {\mathbf s}} p({\mathbf y}, {\mathbf s} | {\mathbf x}){\rm log}
\frac{p({\mathbf s} |{\mathbf x}, {\mathbf y})}{\sum_{{\mathbf y}}p({\mathbf s} | {\mathbf x}, {\mathbf y})p({\mathbf y}| {\mathbf x})}\right]}}
\end{eqnarray}

Intuitively, it measures how much linguistic explanation ${\mathbf s}$ becomes useful information to identify a category ${\mathbf y}$ when given a set of example-based explanation ${\mathcal D}$.

The direct estimation of (\ref{eq:mi1}) is difficult, as calculating the expectation over all possible ${\mathcal D}$ is intractable. 
We handle this problem by (a) introducing the variational lower bound and (b) leveraging reparameterization trick similar to~\cite{chen2018learning}, which are described in subsections~\ref{subsec:vl} and \ref{subsec:sampling}, respectively.

\subsection{Maximizing variational bound}\label{subsec:vl}
In this subsection, we consider the variational lower bound of (A) in (\ref{eq:mi1}). From the definition of the KL divergence, $p\ {\rm log}\ p \ge p\ {\rm log} \  q$ is applied for any distribution $p$ and $q$. Using this relation, (A) inside the expectation in (\ref{eq:mi1}) can be lower-bounded as follows:
\begin{eqnarray}\label{eq:lowerbound}
({\rm A}) \ge \sum_{{\mathbf y}, {\mathbf s}, {\mathcal D}} p({\mathbf s}, {\mathcal D} | {\mathbf x}, {\mathbf y})p({\mathbf y} |  {\mathbf x}) {\rm log} \frac{q({\mathbf s} | {\mathbf x}, {\mathbf y}, {\mathcal D})}{p({\mathbf s} |{\mathbf x}, {\mathcal D})}
\end{eqnarray}
$q({\mathbf s} | {\mathbf x}, {\mathbf y}, {\mathcal D})$ can be any distribution provided that it is normalized, and the expectation of the KL divergence between $q({\mathbf s} | {\mathbf x}, {\mathbf y}, {\mathcal D})$ and true distribution $p({\mathbf s} | {\mathbf x}, {\mathbf y}, {\mathcal D})$ is the difference between (A) and the lower bound.
Similar to the method in~\cite{gao2016variational}, we used the following
\begin{eqnarray}
q({\mathbf s}|{\mathbf x}, {\mathbf y}, {\mathcal D})  =
\frac{q({\mathbf s}, {\mathbf y} | {\mathbf x}, {\mathcal D})}{q({\mathbf y} | {\mathbf x}, {\mathcal D})} 
=
\frac{q({\mathbf y} | {\mathbf x}, {\mathbf s}, {\mathcal D})p({\mathbf s} | {\mathbf x}, {\mathcal D})}{\sum_{s'} q({\mathbf y} |{\mathbf x}, {\mathbf s}', {\mathcal D}) p({\mathbf s}' | {\mathbf x}, {\mathcal D})}, \nonumber
\end{eqnarray}
and substituted it to (\ref{eq:lowerbound}).
When considering parameterizing as in (\ref{eq:decomp}), it is computationally difficult to calculate $p({\mathbf s} | {\mathbf x}, {\mathcal D})$.  
Considering the sampling order, we approximate it to $p({\mathbf s} | {\mathbf x}, {\mathbf y})$ instead for simplicity. 
The first term of the objective function used for optimization is as follows:
\begin{eqnarray}\label{eq:lowerbound2}
(\ref{eq:lowerbound}) \approx {\mathbb E}_{p({\mathbf y}, {\mathbf s}, {\mathcal D}| {\mathbf x})}\left[{\rm log}\ \frac{q({\mathbf y} | {\mathbf x}, {\mathbf s}, {\mathcal D})}{{\mathbb E}_{p({\mathbf s} | {\mathbf x}, {\mathbf y})}[q({\mathbf y} | {\mathbf x}, {\mathbf s}, {\mathcal D})]}\right].
\end{eqnarray}

$q({\mathbf y} | {\mathbf x}, {\mathbf s}, {\mathcal D})$ is hereafter referred to as a reasoner, which is expected to {\it reason} the category of input given a pair of explanation for it.

\subsection{Continuous relaxation of subset sampling}\label{subsec:sampling}
The abovementioned (\ref{eq:lowerbound2}) is required to be optimized stochastically with sampling to avoid calculating the summation over the enormous number of possible combinations of ${\mathcal D}$. 
In this situation, the difficulty of optimization with regard to the network parameter still exists. As it involves the expectation over the distribution to be optimized, 
sampling process disables calculating the gradient of parameters, rendering it impossible to apply back-propagation.

We resort on the reparameterization trick to overcome this issue, which replaces the non-differential sampling process to the deterministic estimation of the distribution parameter, followed by adding random noise. In particular, the Gumbel-softmax~\cite{maddison2016concrete, jang2016categorical} function is utilized similar to ~\cite{chen2018learning}, which approximates a random variable represented as a one-hot vector sampled from a categorical distribution to a vector using continuous values. 
Specifically, we estimate the parameter of categorical distribution ${\mathbf p} \in R^{N}$ satisfying $\sum^{N}_{i=1}  p_i=1$ by the network where $N=|{\mathcal X}|$ is the candidate set of examples. An $N$-dimensional vector ${\mathbf C}$,  which is a continuous approximation of the categorical one-hot vector, is sampled by applying softmax to the estimated parameter after taking logarithm and adding a noise ${\mathbf G}$ sampled from the Gumbel distribution as follows:
\begin{equation}
{\mathbf C}[i] = \frac{{\rm exp}\{({\rm log}\ p_i + G_i)/\tau\}}{\sum^{N}_{j=1} {\rm exp}\{({\rm log}\ p_j + G_j)/\tau\}}
\end{equation}
where 
\begin{equation}
G_i = -{\rm log}(-{\log u_i}), u_i \sim {\rm Uniform}(0, 1),
\end{equation}
and $\tau$ is the temperature of softmax controlling the hardness of the approximation to the discrete vector.
To sample $k$-hot vector representing example-based explanation ${\mathcal D}$, concrete vector ${\mathbf C}$ is independently sampled $k$ times, and  element-wise maximum is taken to ${\mathbf C}_1, {\mathbf C}_2, \ldots, {\mathbf C}_k$ to construct a vector corresponding to ${\mathcal D}$.

\subsection{Structure of networks}\label{subsec:structure}
We parameterize three probabilistic distributions, explainer, selector, and reasoner with different neural networks. We elucidate their detailed structures.

\begin{description}[labelindent=0.2cm, leftmargin=0cm]
\item[Explainer $p({\mathbf s} | {\mathbf x}, {\mathbf y})$] is represented by a neural network that predicts the probability of each type (dimension) of attribute is selected.
The model is constituted using three fully-connected layers as in left of Fig.~\ref{fig:exp}. Taking the target sample ${\mathbf x}$ and the category label ${\mathbf y}$ as inputs, the model projects them to the common-space and element-wise summation is applied. After one more projection, they are normalized by the the softmax function. 
The output dimension of the network ${\mathbf f}({\mathbf x}, {\mathbf y})$ is the same as that of the attribute vector, and each dimension indicates the probability that each type of attribute is selected as an explanation.  
When training, the attribute ${\hat {\mathbf s}}$ assigned to the sample is used as the value. Formally, for all $i$-th dimension of the linguistic explanation vector,
\[
p({\mathbf s}|{\mathbf x}, {\mathbf y}) = 
\begin{cases}
  {\mathbf f}({\mathbf x}, {\mathbf y})[i] & \text{if ${\mathbf s}[i]$\ =\ ${\hat {\mathbf s}}[i]$} \\
  0 & \text{otherwise}
\end{cases}
\]\\
For inference, the value that maximizes the output of the reasoner (described later) for the class to be explained is selected.

\item[Selector $p({\mathcal D}| {\mathbf x},  {\mathbf y}, {\mathbf s})$] takes the linguistic explanation ${\mathbf s}$ in addition to ${\mathbf x}$ and ${\mathbf y}$ as inputs; their element-wise summation is calculated after projecting them to the common-space. As stated in the previous subsection, we leverage reparameterization trick to render the optimization tractable owing to the enormous number of the combination ${\mathcal D}$. 
The network estimates the parameter ${\mathbf p}$ of categorical distribution. When sampling from a distribution, noise variables that are independently generated $k$ times are added to the parameter, and the element-wise maximum is computed after the Gumbel softmax is applied.
\item[Reasoner $q({\mathbf y} | {\mathbf x}, {\mathbf s}, {\mathcal D})$] 
infers the category to which the sample ${\mathbf x}$ belongs, given a pair of generated explanation (${\mathbf s}$, ${\mathcal D}$). 
We design it by modifying the matching network~\cite{vinyals2016matching}, which is a standard example-based classification model. 
The prediction of the reasoner must be based on the given explanations. Such reasoning process is realized by considering (1) consistency to the linguistic explanation ${\mathbf s}$, and (2) similarity to the target sample ${\mathbf x}$, for each example in ${\mathcal D}$. 

Based on a certain reason, the reasoner decides whether each example deserves consideration, and predicts the category exploiting only selected examples. The weight of each referred sample ${\mathbf x}_i$ is determined by the visual and semantic similarity to the target ${\mathbf x}$. 
More formally,
\begin{eqnarray}\label{eq:matching}
q({\mathbf y} | {\mathbf x}, {\mathbf s}, {\mathcal D})= \sum_{( {\mathbf x}_i, {\hat {\mathbf s}_i}, {\mathbf y}_i) \in {\mathcal D}} \alpha({\mathbf s}, {\hat {\mathbf s}_i)}\ w({\mathbf x}, {\mathbf s}, {\mathbf x}_i, {\hat {\mathbf s}_i}) \ {\mathbf y}_i  \label{eq:1}\\
q({\bar y} | {\mathbf x}, {\mathbf s}, {\mathcal D}) =  1 - \sum_{( {\mathbf x}_i, {\hat {\mathbf s}_i}, {\mathbf y}_i) \in {\mathcal D}} \alpha({\mathbf s}, {\hat {\mathbf s}_i)}\ w({\mathbf x}, {\mathbf s}, {\mathbf x}_i, {\hat {\mathbf s}_i})
\label{eq:2}
\end{eqnarray}
where 
\begin{eqnarray}\label{eq:weight}
w({\mathbf x}, {\mathbf s}, {\mathbf x}_i, {\hat {\mathbf s}_i}) = \frac{{\rm exp}({\mathbf g}({\mathbf x}, {\mathbf s})^{\top} {\mathbf g}({\mathbf x}_i, {\hat {\mathbf s}_i)})} {\sum\limits_{({\mathbf x}_i, {\hat {\mathbf s}_i}, {\mathbf y}_i) \in {\mathcal D}}^{} \!\!\! {\rm exp}({\mathbf g}({\mathbf x}, {\mathbf s})^{\top} {\mathbf g}({\mathbf x}_i, {\hat {\mathbf s}_i)})} 
\end{eqnarray}
$\alpha$ indicates the function to verify the coincidence of the linguistic explanation and the attribute assigned to each sample. In our setting, we set as $\alpha({\mathbf s}, {\hat {\mathbf s}}) = \sum_i [[{\mathbf s}[i] = {\hat {\mathbf s}}[i]]]$ where $[[\cdot]]$ is an indicator function of 1 if the condition inside bracket are satisfied, otherwise 0. Note $\alpha({\mathbf s}, {\hat {\mathbf s}}) \in \{0, 1\}$ as ${||\mathbf s}||_{0}=1$.
$w$ measures the weight of each referred sample used for prediction. 
The probability of the sample being assigned to each class is determined to utilize the samples in $\mathcal D$, which match to the linguistic explanation as in (\ref{eq:1}). An additional ``unknown'' category ${\bar y}$ is introduced for convenience, indicating the inability to predict from the input explanations. The remaining weight is assigned to the probability of the ``unknown'' class, as in (\ref{eq:2}). In (\ref{eq:weight}), ${\mathbf g}({\mathbf x}, {\mathbf s})$ is the feature embedding implemented 
by the neural network as in the right-most in Fig.~\ref{fig:structure}, and the similarity is computed by the dot product in that space following normalization by the softmax function.

While the reasoner attempts to make a decision based on the given explanations, the other two models are trained collaboratively to generate explanations such that the reasoner can reach the appropriate conclusion. 
\end{description}

\subsection{Training and Inference}\label{subsec:optimize}
We parameterize the joint distribution as in (\ref{eq:decomp}), and the lower bound of the objective (\ref{eq:interaction}) calculated by (\ref{eq:mi2}) and (\ref{eq:lowerbound2}) is optimized with regard to the parameters of the neural network models representing $p({\mathbf s | {\mathbf x}, {\mathbf y}})$, $p({\mathcal D} | {\mathbf x}, {\mathbf y}, {\mathbf s})$, and $q({\mathbf y} | {\mathbf x}, {\mathbf s}, {\mathcal D})$. Assuming that the calculation of the expectation over ${\mathbf s}$ is feasible, although that over ${\mathcal D}$ is not, we optimized the model of the selector by sampling, and that of the explainer was optimized directly.

The processing flow in each iteration is as follows:
\begin{enumerate}[noitemsep,topsep=0pt, leftmargin=5.5mm]
\setlength{\itemsep}{0pt}
  \setlength{\parskip}{0pt}
\item ${\mathbf x}$ is sampled randomly from the training dataset,
\item ${\mathbf y}$ is sampled randomly from the predictor $p({\mathbf y} | {\mathbf x})$,
\item $p({\mathbf s} | {\mathbf x}, {\mathbf y})$ is computed for possible ${\mathbf s}$,
\item ${\mathcal D}$ is sampled randomly from the selector $p({\mathcal D} | {\mathbf x}, {\mathbf y}, {\mathbf s})$ for each ${\mathbf s}$,
\item For each sampled $({\mathbf x}, {\mathbf y}, {\mathbf s}, {\mathcal D})$, the objective is calculated by (\ref{eq:lowerbound2}) and (\ref{eq:mi2}), and the gradient of it w.r.t the weights of all parametric models are computed. 
\item All weights are updated by stochastic gradient decent (SGD).
\end{enumerate}

The inference is performed by sequentially sampling variables from the distributions given input ${\mathbf x}$. When generating linguistic explanations, $M$ identical attribute type is selected whose output value of the predictor is the largest, where $M$ is the number of attribute-examples pairs. used for explanation. For estimating the attribute value, the one that most explains the prediction the best will be selected. In other words, ${\mathbf s}_1$, ${\mathbf s}_2, ...$ having the same attribute type, the value maximizing $q({\mathbf y}|{\mathbf x}, {\mathbf s}, {\mathcal D})$ is outputted after the corresponding ${\mathcal D}_1, {\mathcal D}_2, ...$ are sampled from the selector. 

\begin{figure}[t!]
  \includegraphics[width=0.95\linewidth]{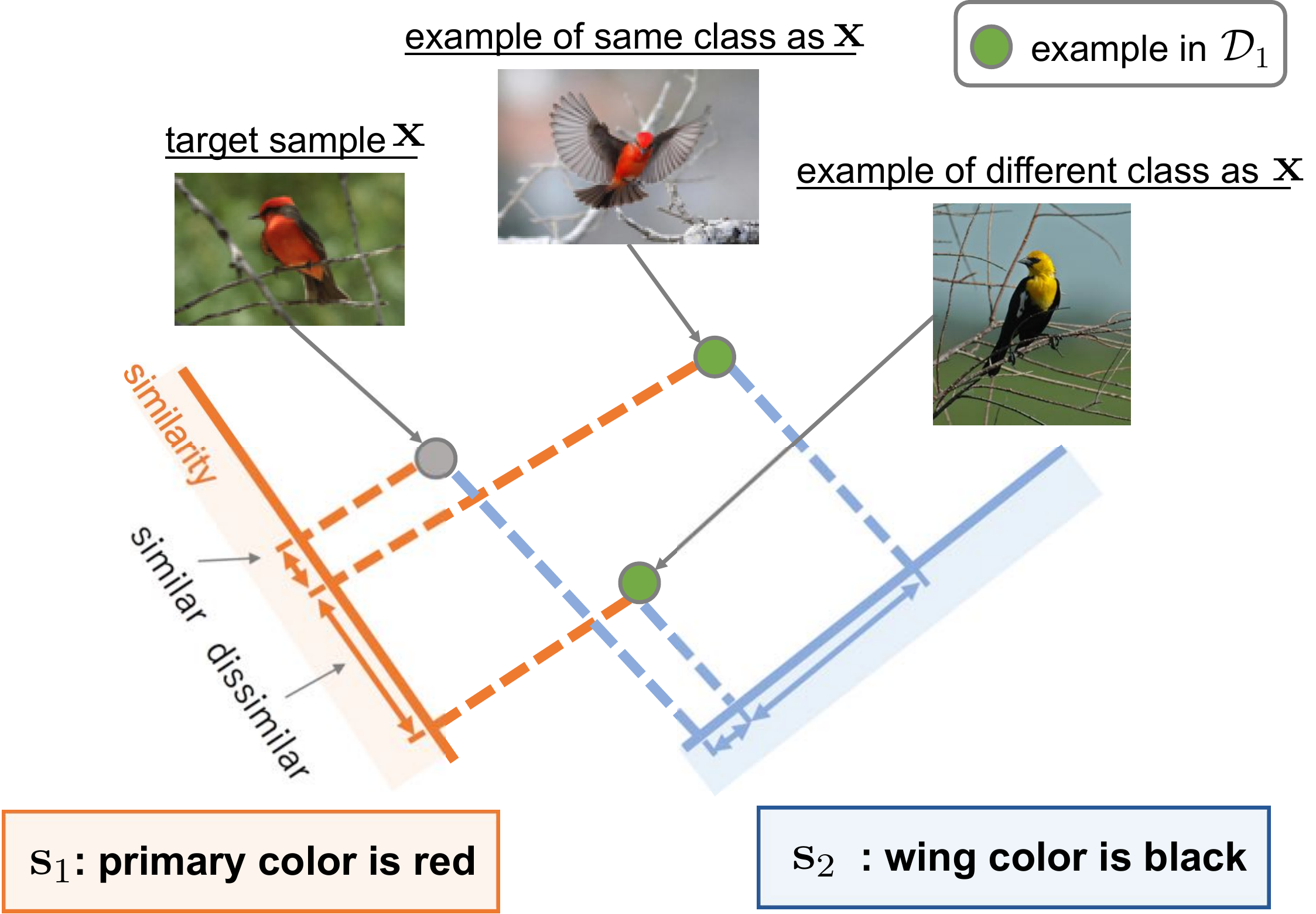}
  \vspace{-0.1cm}
  \caption{ Intuitive understanding of complemental explanations. The reasoner predicts the target sample ${\mathbf x}$ (written as gray circles) by referring other samples based on the similarity space (orange and blue) corresponding to each linguistic explanation ${\mathbf s}_1, {\mathbf s}_2$. Considering two pairs of possible explanations $({\mathbf s}_1, {\mathcal D}_1)$ and $({\mathbf s}_2, {\mathcal D}_2)$, the expected ${\mathcal D}_1$ (written as green circle) is the one by which the reasoner can reach the correct conclusion with ${\mathbf s}_1$; however, this cannot be achieved with ${\mathbf s}_2$.}
  \label{fig:intexp}
  \vspace{-0.3cm}
\end{figure}

\subsection{Which explanation is complemental?}\label{subsec:discuss}
By analyzing the proposed method, it provides an intuitive understanding of complemental explanations, from the viewpoint of maximizing interaction information.

To understand which set ${\mathcal D}$ is preferred, we consider (\ref{eq:lowerbound2}) where ${\mathcal D}$ relates. Inside the expectation in this equation, the numerator is the output of the reasoner, and the denominator is that averaged over ${\mathbf s'}$. Given ${\mathbf x}$, ${\mathbf y}$, and ${\mathbf s}$, the situation where the ratio becomes large is when the reasoner can reach the correct conclusion ${\mathbf y}$ for given linguistic explanation ${\mathbf s}$ with ${\mathcal D}$ but it can not when ${\mathcal D}$ is used with other ${\mathbf s'}$.
In other words, an example-based explanation is complemental to its linguistic counterpart when it is a {\it discriminative set} of examples for not only the target but also the linguistic explanation.

The concept of ``a set is discriminative'' is clearly different from ``single example is discriminative'' in our framework. This can be understood intuitively by Fig.~\ref{fig:intexp}. A reasoner contains a different similarity space for each linguistic explanation ${\mathbf s}$.
Here, we consider two possible explanations ${\mathbf s}_1, {\mathbf s}_2$, and ${\mathcal D}_1$ which is the counterpart of ${\mathbf s}_1$.
In this situation, the desired ${\mathcal D}$ for linguistic explanation ${\mathbf s}$ is that the correct class is predicted for the given ${\mathbf s}$, but a wrong one is predicted for different ${\mathbf s}'$.
Therefore, the set should contain both of the same/different classes from the predicted one. As shown, a naive method of example selection, such as one selecting only one nearest sample from the target, is not appropriate for selecting a complemental explanation.

Considering ${\mathbf s}$, it relates to both terms in (\ref{eq:interaction}). For (\ref{eq:lowerbound2}), the same claim as that mentioned above can be applied: a complemental linguistic explanation ${\mathbf s}$ for examples ${\mathcal D}$ is one where a specific set ${\mathcal D}$ can be derived, instead of another see ${\mathcal D}'$.
As for (\ref{eq:mi2}), 
it can be regarded as a regularizer to avoid weighing excessively on the attribute that can identify the target class without considering examples for selecting${\mathbf s}$.

\subsection{Relationship with other methods}\label{subsec:relation}
The existing works for visual explanation
explanations (e.g.,~\cite{hendricks2016generating})
trains the classifier as well as the explanation generator to guarantee that the generated explanations are discriminative for the target class. 
In this work, we also train the auxiliary classifier (i.e., reasoner) similar to the existing methods; however, it naturally appears in the context of interaction information (mutual information) maximization. Conversely, we found that such an intuitive idea in these works is justified from the information theoretical viewpoint. Similarly, our method shares the idea with methods for generating referring expression (e.g.,\cite{yu2017joint}) in that they utilize auxiliary models.

\section{Experiment}\label{sec:experiment}
We conducted experiments to verify that the proposed method can generate the appropriate explanation.
Given a target sample ${\mathbf x}$, our system generates a prediction ${\mathbf y}$ from the predictor, and explanations $({\mathbf s}_1, {\mathcal D}_1)$, $({\mathbf s}_2, {\mathcal D}_2)$, $...$, $({\mathbf s}_M, {\mathcal D}_M)$ from explanation models. 
We evaluated the proposed method by quantifying the properties that the generated explanation should satisfy: (a) fidelity and (b) complementarity. 
Related to (a), we consider two types of fidelity as follows. 
(a1) The target value ${\mathbf y}$ to be explained should be obtained from the explanations. Moreover, (a2) The outputted linguistic explanation ${\mathbf s}$ should be correct. 
In addition, as for (b), we would like to assess whether the output explanations $({\mathbf s}, {\mathcal D})$ are complemental to each other. 
In the following subsections, we describe the evaluation method as well as discussing the obtained results after elucidating the setting of the experiments in subsection~\ref{subsec:setting}.

\begin{table}[t!]
\small 
  \centering
   \begin{tabular}{|c||c|c|c|c|} \hline 
    dataset & acc (predictor) & acc (reasoner) & consistency   \\ \hline\hline 
    AADB & 0.647 &  0.646 & 0.738  \\ \hline 
    CUB & 0.694 &  0.434 & 0.598  \\ \hline 
   \end{tabular}
     \vspace{-0.2cm}
     \caption{The accuracy of identifying the target category of the predictor (target) and reasoner (explain), and the consistency between them.}
  \label{tab:consistency}
   \vspace{-0.4cm}
\end{table}

\begin{table}[t!]
\small 
  \centering
   \begin{tabular}{|c||c|c|c|c|} \hline 
    dataset & baseline (random) & baseline (predict) & ours   \\ \hline\hline 
    AADB & 0.200 & 0.572 &  0.582  \\ \hline 
    CUB & 0.125 & 0.428 &  0.436 \\ \hline 
   \end{tabular}
     \vspace{-0.1cm}
     \caption{The accuracy of identifying the attribute value of our model and that of baselines: selecting attribute value randomly (random), and  predicting attributes by the perceptron (predict).}
       \label{tab:faithful}
    \vspace{-0.6cm}
\end{table}

\subsection{Experimental setting}\label{subsec:setting}
\textbf{Dataset}
In our experiments, we utilized Caltech-UCSD Birds-200-2011 Dataset (CUB)~\cite{WahCUB_200_2011} and Aesthetics with Attributes Database (AADB)~\cite{kong2016photo}, both of which hold attributes assigned for all contained images. CUB is a standard dataset for fine-grained image recognition, and it contains 11,788 images in total and 200 categories of bird species. It contains 27 types of attributes, such as ``wing pattern'' or ``throat color.'' 
AADB is a dataset created for the automatic image aesthetics rating. It contains 10,000 images in total and the aesthetic score in [-1.0, 1.0] is assigned for each image. We treat the binary classification by regarding images having non-negative scores as samples of the positive class, and remaining samples as the negative class. Attributes are also assigned as the continuous values in [-1.0, 1.0], and we discretized them according to the range that it belongs to: [-1.0, -0.4), [-0.4, -0.2), [-0.2, 0.2), [0.2, 0.4), or [0.4, 1.0]. It contains eleven types of attributes, including ``color harmony'' and ``symmetry.''
Unlike the standard split, we utilized 60\% of the test set for CUB, and 30\% of the train set for AADB as the candidates of examples ${\mathcal X}$.

Although CUB dataset is for the fine-grained task, where the inner-class variance of the appearance is considered small, that of AADB is large owing to the subjective nature of the task. We selected these two datasets to assess the influence of the variation of the samples within the class.

\textbf{Detailed setting}
To prepare a predictor, we fine-tuned a deep residual network~\cite{he2016deep} having 18 layers for each dataset, which is pre-trained on ImageNet dataset~\cite{deng2009imagenet}. The optimization was performed with SGD. The learning rate, weight decay, momentum, and batch size were set to 0.01, $10^{-4}$, 0.9, and 64, respectively. When training explanation models, all networks were optimized with SGD without momentum with learning rate $10^{-3}$, weight decay $10^{-3}$, and batch size 64 for AADB and 20 for CUB. We set $k$, the number of examples used for explanations, to 10 in all experiments. 

Empirically, we found that the linguistic explainer 
$p({\mathbf s} | {\mathbf x}, {\mathbf y})$ tended to assign a high probability (almost 1) on only one type of attribute, and small probability (almost 0) to the others.  
To avoid it, we added an extra entropy term $H({\mathbf s}|{\mathbf x}, {\mathbf y}) = \sum_{\mathbf s}-p({\mathbf s}| {\mathbf x}, {\mathbf y})\ {\rm log}\ p({\mathbf s}| {\mathbf x}, {\mathbf y})$ to the maximized objective function as our goal is to generate multiple outputs.
The implementation was performed on the Pytorch framework~\cite{paszke2017automatic}. 

\begin{table}[t!]
\small 
  \centering
   \begin{tabular}{|c||c|c|c|c|} \hline 
        & ours & w/o ${\mathbf x}$ & w/o ${\mathbf y}$ & w/o ${\mathbf s}$   \\ \hline\hline
    accuracy &0.646 & 0.627 &  0.569 & 0.613  \\ \hline 
    consistency & 0.738 & 0.689 &  0.600 & 0.620  \\ \hline 
   \end{tabular}\\
   \vspace{0.10cm}
   \begin{tabular}{|c||c|c|c|c|} \hline 
        & ours & w/o ${\mathbf x}$ & w/o ${\mathbf y}$ & w/o ${\mathbf s}$   \\ \hline\hline
    accuracy &0.434 & 0.354 &  0.02 & 0.153  \\ \hline 
    consistency & 0.598 &0.492 &  0.02 & 0.201  \\ \hline 
   \end{tabular}
     \vspace{-0.2cm}
 \caption{The ablation study for the accuracy of identifying the target category on AADB dataset (above) and CUB dataset (below).}
   \label{tab:abl_CUB}
  \vspace{-0.7cm}
\end{table}

\subsection{Fidelity} 
One important factor of the explanation model is the fidelity to the target to be explained.
We conducted an experiment to investigate whether the target can be obtained from its explanation. Interestingly, our framework holds two types of paths to the decision. One is the target predictor $p({\mathbf y}|{\mathbf x})$ to be explained.  
The other is the route via explanations interpretable for humans, i.e.,
${\mathbf y}\rightarrow{} {\mathbf s}, {\mathcal D} \rightarrow{}{} {\mathbf y}'$ through the explainer $p({\mathbf s} | {\mathbf x}, {\mathbf y})$, selector $p({\mathcal D} | {\mathbf x}, {\mathbf y}, {\mathbf s})$, and reasoner $q({\mathbf y}' | {\mathbf x}, {\mathbf s}, {\mathcal D})$. 
We evaluated the fidelity to the model by the consistency between the interpretable decisions from the latter process and that from the target predictor. In the Table.~\ref{tab:consistency}, we reports the consistency as well as the mean accuracy of each models. As shown, the explanation model (written as reasoner) achieved the similar performance as the target model (written as predictor), and considerably high consistency on both datasets.

We also conducted the ablation study to clarify the influence of three variables ${\mathbf x}, {\mathbf y}, {\mathbf s}$ on the quality of explanations. We measured the accuracy in the same manner as above except that we dropped one variable by replacing the vector filled by 0 when generating explanations. The results in Table~\ref{tab:abl_CUB} exhibits that our models put the most importance on the category label out of three. These results are reasonable because it contains information for which the explanation should be discriminative.

The other important aspect for the fidelity in our task is the correctness of the linguistic explanation.
In particular, the attribute value (e.g., ``red'' or ``blue'' for attribute type ``color'') is also estimated during the inference. We evaluated the validity by comparing the predicted attributes with that of grand-truth on the test set. 
The attribute value that explains the output of the predictor ${\mathbf y}$ the best will be selected as written in subsection~\ref{subsec:optimize}. 
As baselines, we employed the three layers perceptron with a hidden layer of 512 dimensions (predict). It was separately trained for each type of attribute with SGD. Moreover, we also report the performance when the attribute is randomly selected (random).  
We measured the accuracy and the results are shown in the Table~\ref{tab:faithful}. As shown, our method, which generates linguistic explanations through selecting examples, can predict it as accurate as the direct estimation, and the accuracy is much better than the random selection.

\begin{figure}[t!]
\centering
\begin{minipage}{.48\linewidth}
  \centering
 \includegraphics[width=\linewidth, height=5cm]{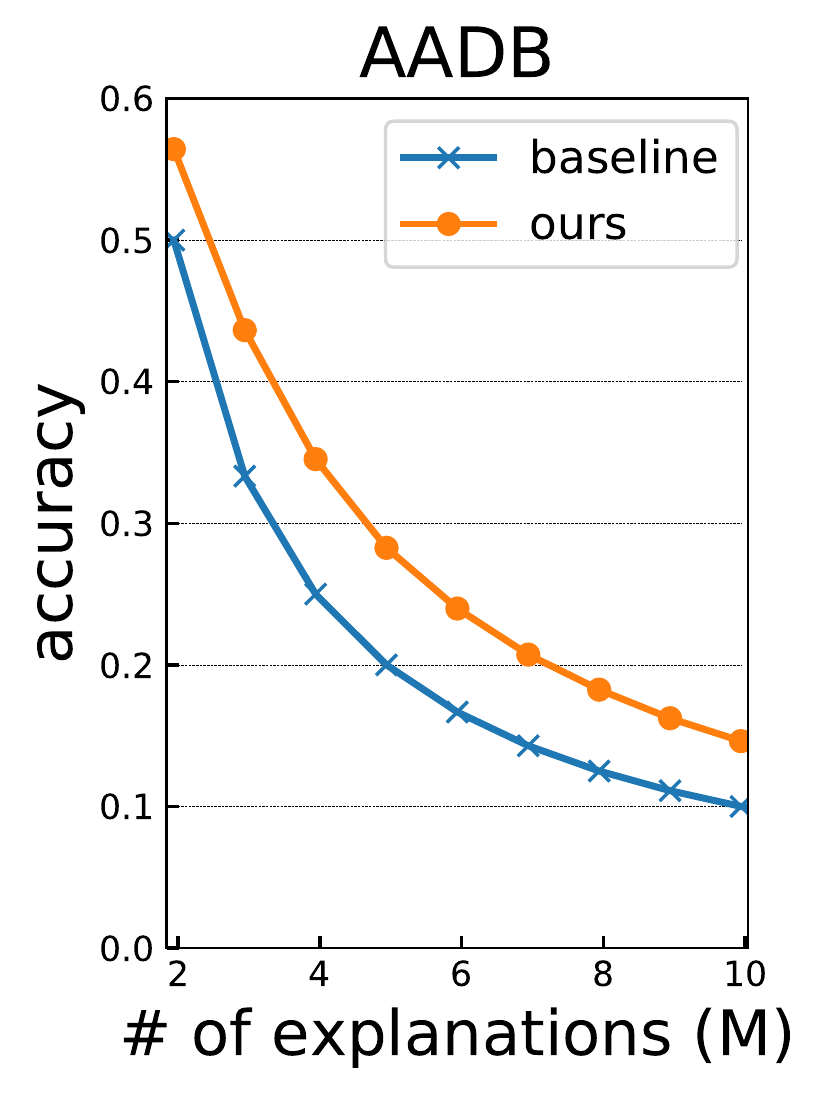}
 \end{minipage}
\begin{minipage}{.48\linewidth}
  \centering
  \includegraphics[width=\linewidth, height=5cm]{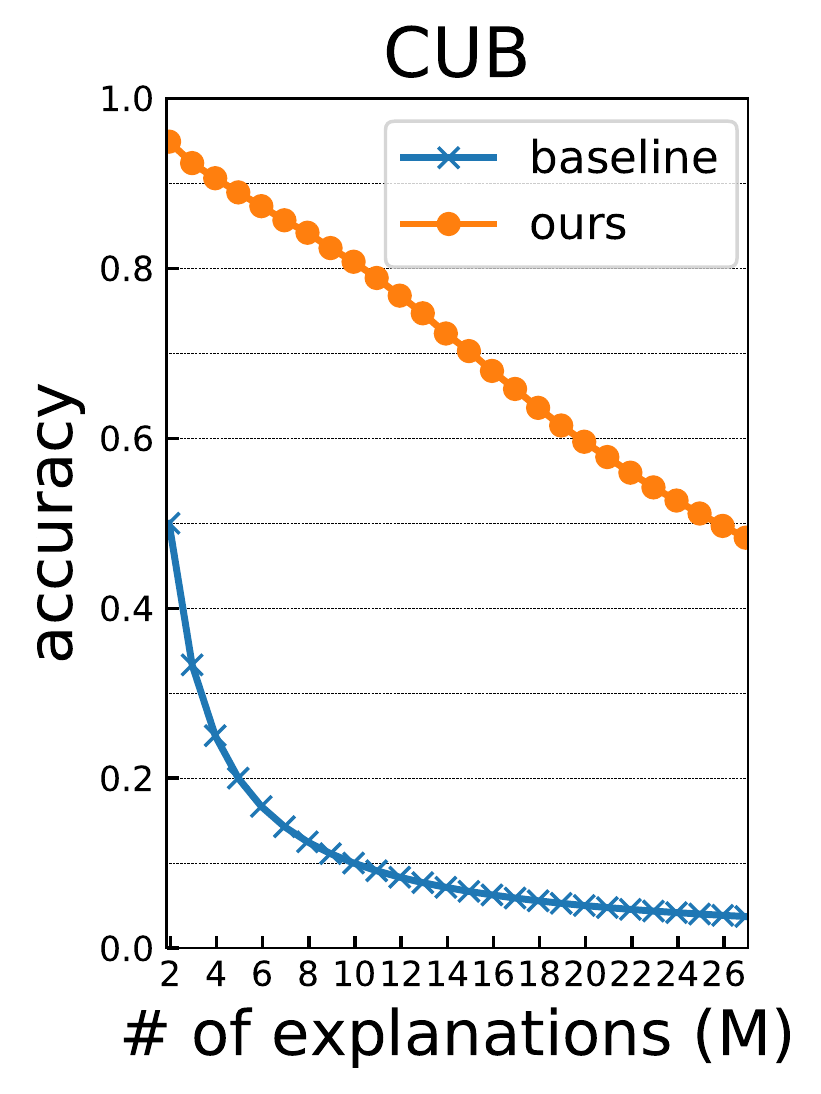}
\end{minipage}
\vspace{-0.1cm}
\caption{The mean accuracy of identifying the linguistic explanation from the examples on AADB (left) and CUB (right) dataset. The y-axis and x-axis indicates the accuracy and the number of generated explnations.}
\label{fig:curve}
\vspace{-0.5cm}
\end{figure}

\subsection{Complementarity}
To quantify the complementarity of explanations, we investigate how the example-based explanation ${\mathcal D}$ renders the linguistic explanation ${\mathbf s}$ identifiable. 
Specifically, utilizing the reasoner $q({\mathbf y}|{\mathbf x}, {\mathbf s}, {\mathcal D})$, which is trained to reason the target from the explanation, we confirmed whether it can reason to the correct conclusion only from the generated explanation pair as discussed in subsection~\ref{subsec:discuss}.
For generated pairs of explanations $({\mathbf s}_1, {\mathcal D}_1)$, $({\mathbf s}_2, {\mathcal D}_2), ...$, $({\mathbf s}_M, {\mathcal D}_M)$ of which attribute type is identical, we computed the output of the reasoner as $q_{ij} = q({\mathbf y}|{\mathbf s}_i, {\mathcal D}_j)\ (1\le i, j\le M$) for ${\mathbf y}$ obtained from the predictor. 
Selecting the index having the maximum value for all $j$ as $i^{\star} = \argmax_i q_{ij}$, we verified $i^{\star}=j$. The mean accuracy is compared with a baseline that outputs the same examples for all ${\mathbf s}_i$ and results are shown in Fig.~\ref{fig:curve}. The x-axis of the figure indicates the number of the generated explanations (i.e., M). 

\begin{figure}[t!]
\centering
\begin{minipage}{.48\linewidth}
  \centering
 \includegraphics[width=\linewidth, height=3.5cm]{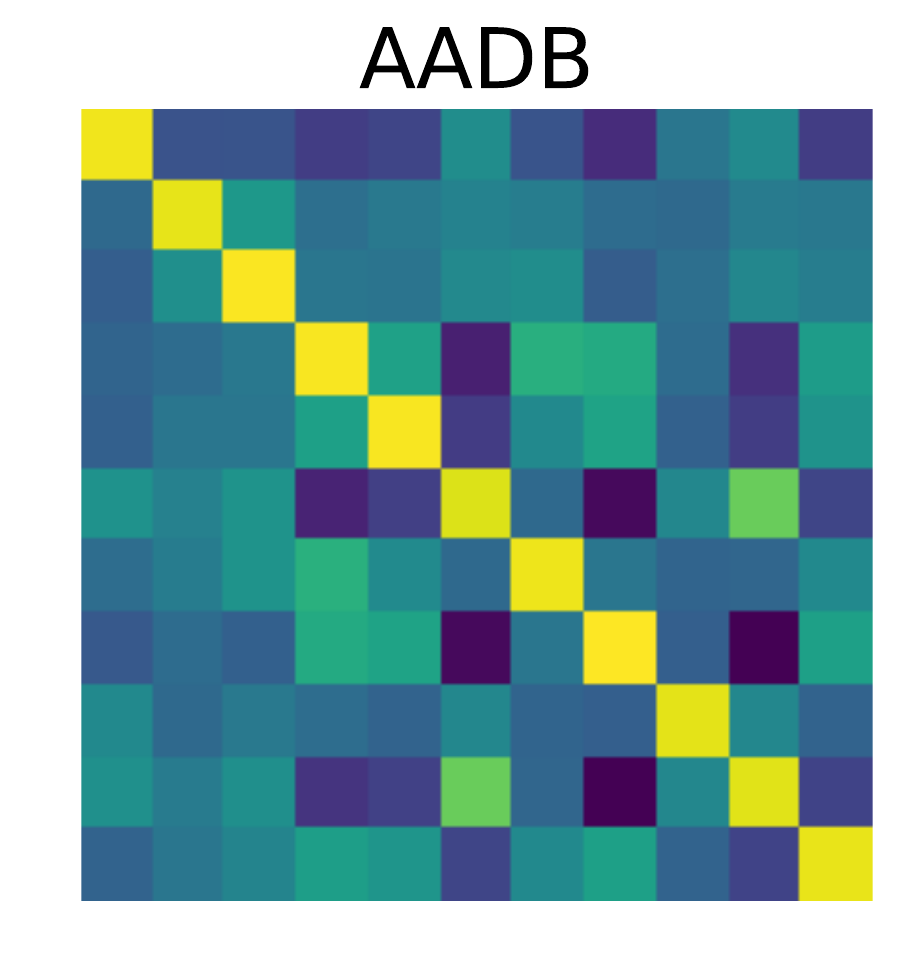}
 \end{minipage}
\begin{minipage}{.48\linewidth}
  \centering
  \includegraphics[width=\linewidth, height=3.5cm]{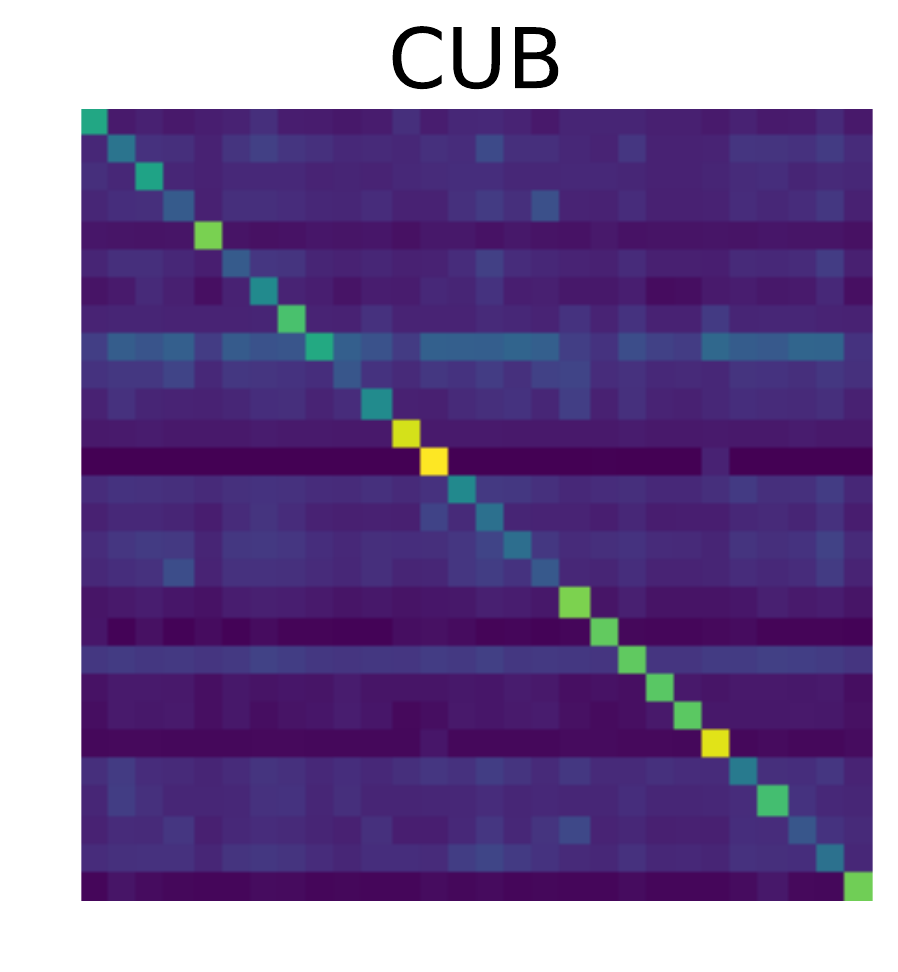}
\end{minipage}
\vspace{-0.3cm}
\caption{The confusion matrix of identifying the attribute type from the examples on AADB (left) and CUB (right) dataset.}
\label{fig:confusion}
\vspace{-0.3cm}
\end{figure}

On both datasets, the accuracy of our model is better than the baseline.
Furthermore, as shown in Fig.~\ref{fig:confusion}, we observe that the diagonal element has a high value on the confusion matrix.
These results demonstrate the ability of our method to generate complemental explanations. 
The difference in the performance between the proposed method and baseline on AADB is lower than on CUB. 
We conjecture that one reason is the difference between appearance and attributes. AADB dataset contains highly-semantic attributes (e.g., ``color harmony'') compared with those in CUB (e.g., ``color'' or ``shape''). Such the semantic gap may hinder to construct the discriminative set which renders the attribute identifiable.

\begin{figure}[t!]
\vspace{-0.2cm}
  \includegraphics[height=3.9cm]{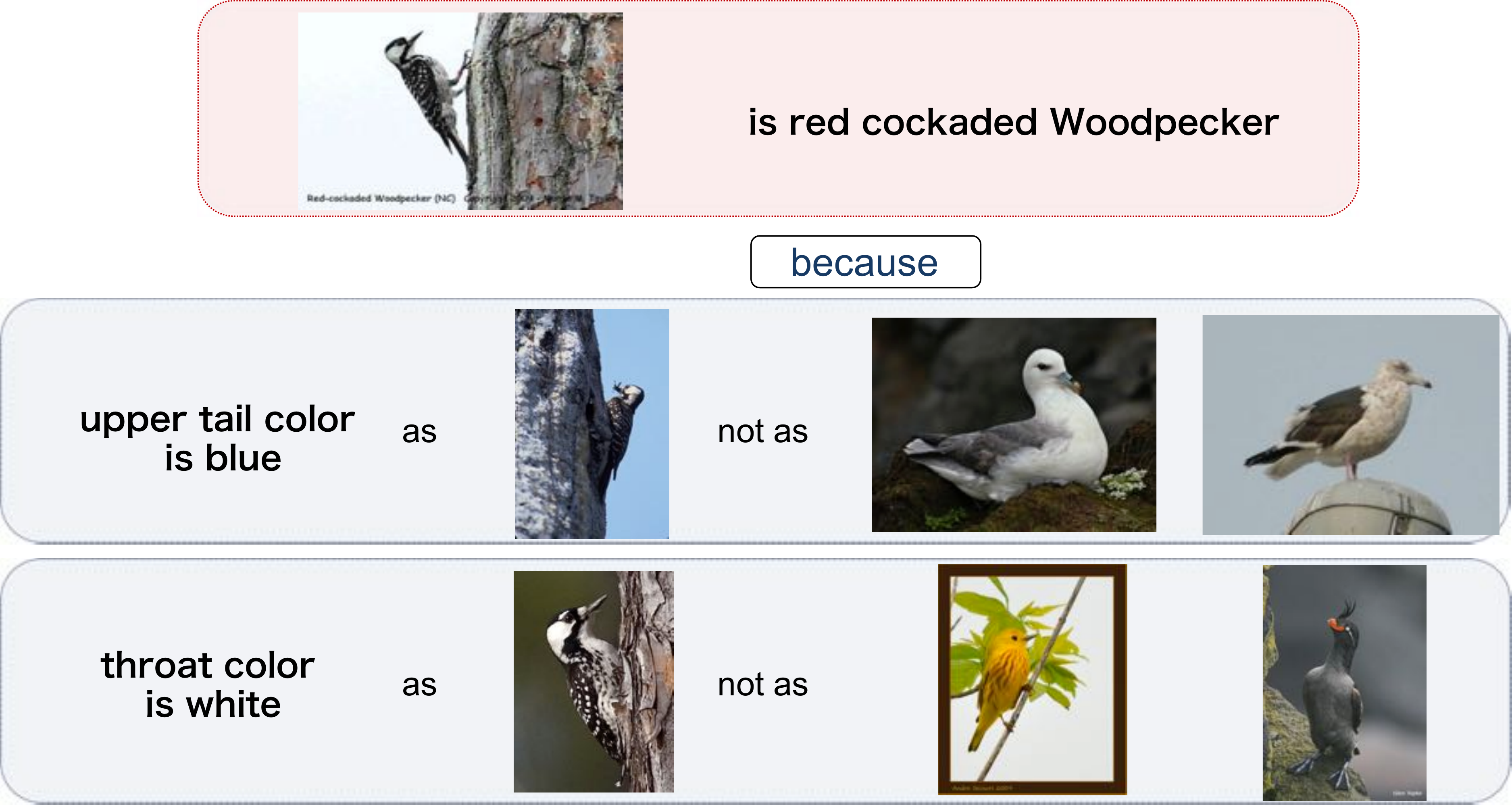}
  \caption{Example output of our system on CUB dataset.}
  \vspace{-0.7cm}
  \label{fig:example_output}
\end{figure}

\subsection{Output example}
An output of our system on CUB dataset when the number of explanations is two is shown in Fig.~\ref{fig:example_output}. 
In this example, the combination of linguistic and example-based explanation seems compatible, where it will not make sense if these pairs are switched. For instance, the below linguistic explanation ``throat color is white'' may not be a good explanation for the above examples.

Although not the primary scope in this work,
the proposed task may be extended to machine teaching task, where the machine {\it teaches} to human by showing examples iteratively.

\section{Conclusion}\label{sec:conclusion}
In this work, we performed a novel task, that is, generating visual explanations with linguistic and visual examples that are complemental to each other. We proposed to parameterize the joint probability of variables to explain,  and to be explained by the three neural networks. To explicitly treat the complementarity, auxiliary models responsible for the explanations were trained simultaneously to maximize the approximated lower bound of the interaction information. We empirically demonstrated the effectiveness of the method by the experiments conducted on the two visual recognition datasets.

\section{Acknowledgement}
This work was partially supported by JST CREST Grant Number JPMJCR1403, Japan, and partially supported by the Ministry of Education, Culture, Sports, Science and Technology (MEXT) as "Seminal Issue on Post-K Computer." Authors would like to thank Hiroharu Kato, Toshihiko Matsuura for helpful discussions.

{\small
\bibliographystyle{ieee}
\bibliography{egbib}
}

\end{document}